\documentclass[preprint,12pt,authoryear]{elsarticle}
\usepackage{amssymb}
\usepackage{booktabs}
\usepackage{graphicx}
\usepackage{subfigure}

\journal{journal}

\begin{document}

	\begin{frontmatter}
		
		\title{Radious: Unveiling the Enigma of Dental Radiology with BEIT Adaptor and Mask2Former in Semantic Segmentation}

		\author[1]{Mohammad Mashayekhi}
		\ead{mohammad@carestyle.org}
		
		\author[2]{Sara Ahmadi Majd}
		\ead{sara@carestyle.org}
		
		\author[3]{Arian Amiramjadi}
		\ead{arian@carestyle.org}
		
		\author[4]{Babak Mashayekhi}
		\ead{babak@carestyle.org}

		\address[1]{Gilan University of Medical Sciences, Rasht, Iran}
		\address[2]{University of Tehran, Tehran, Iran}
		\address[3]{Hamedan University of Technology, Hamedan, Iran}
		\address[4]{Shahid Beheshti University of Medical Sciences, Tehran, Iran}

		\begin{abstract}
			X-ray images are the first steps for diagnosing and further treating dental problems. So, early diagnosis prevents the development and increase of oral and dental diseases. In this paper, we developed a semantic segmentation algorithm based on BEIT adaptor and Mask2Former to detect and identify teeth, roots, and multiple dental diseases and abnormalities such as pulp chamber, restoration, endodontics, crown, decay, pin, composite, bridge, pulpitis, orthodontics, radicular cyst, periapical cyst, cyst,  implant, and bone graft material in panoramic, periapical, and bitewing X-ray images. We compared the result of our algorithm to two state-of-the-art algorithms in image segmentation named: Deeplabv3 and Segformer on our own data set. We discovered that Radious outperformed those algorithms by increasing the mIoU scores by 9\% and 33\% in Deeplabv3+ and Segformer, respectively.

		\end{abstract}

			
		\end{frontmatter}
		
		
		\section{Introduction}
		Radiological examinations in dentistry assist specialists by displaying the structure of the dental bones to screen embedded teeth, bone abnormalities, cysts, tumors, infections, and fractures 
		\citep{youDeepLearningbasedDental2020}.\\
		Dentists may use X-ray images to examine the complete dental structure for future treatments. X-ray scan is a tool used in dental medicine to examine the condition of a patient's teeth, gums, jaws, and bone structure and to diagnose buccal diseases. The two types of X-rays used in dentistry are intraoral (where the film is placed within the mouth) and extraoral (where the patient’s face is positioned between the radiographic film and the X-ray source). Extraoral panoramic radiography, also known as panoramic X-ray or orthopantomography (OPG), intraoral bitewing radiography, or bitewing X-rays, and periapical intraoral radiography are the three types of dental X-rays.\citep{wangBenchmarkComparisonDental2016}.\\
		Although dentists are responsible for detecting tooth issues, manually analyzing X-ray images can be challenging. For example, subjectivity causes discrepancies in detection among various observers when detecting dental cavities. Furthermore, factors such as radiograph quality, viewing conditions, the dentist's expectations, and the duration of time per examination can all contribute to detection differences \citep{langlaisInterpretationBitewingRadiographs1987b, bailitValidityRadiographicMethod1980a}. Moreover, human error in the manual analysis might result in inaccurate forecasts. Besides, Manual clinical examinations are time-consuming, labor-intensive, and tedious \citep{jaderDeepInstanceSegmentation2018c}.\\
		Researchers have investigated the use of deep learning with convolutional neural networks (CNNs) to analyze many types of medical images in recent years. Deep learning is increasingly used for disease diagnosis, and it has demonstrated precise and expeditious identification with better clinical results \citep{kermanyIdentifyingMedicalDiagnoses2018}.\\
		Another recent approach for analyzing dental images is segmentation. Image segmentation is a field of study that focuses on the problem of grouping pixels in an image. Different semantics for grouping pixels, such as category or instance membership, have led to different types of segmentation tasks, such as panoptic, instance, or semantic segmentation. These tasks differ in semantics, but current methods typically develop specialized architectures for each task \citep{longFullyConvolutionalNetworks2015}.\\
		But using CNNs will induce convolution biases that the model cannot learn global futures and complex relations in images \citep{dosovitskiyImageWorth16x162021a}. Vision Transformer (ViT) is a recent convolution-free transformer architecture for image classification \citep{dosovitskiyImageWorth16x162021}. It processes input images as sequences of patch tokens and requires training on large datasets. The segmenter extended the Vision Transformer for semantic segmentation and built on it \citep{strudelSegmenterTransformerSemantic2021}.\\
		However, the plain Vision Transformer (ViT) has been found to have defects in dense predictions compared to vision-specific transformers. This is due to the lack of image-related prior knowledge, which leads to slower convergence and lower performance. As a result, plain ViTs cannot compete with vision-specific transformers \citep{huangShuffleTransformerRethinking2021}. Researchers have proposed the Vision Transformer Adapter (ViT-Adapter) to address this issue. This pre-training-free network can efficiently adapt the plain ViT to downstream dense prediction tasks without modifying its original architecture. The ViT-Adapter was designed to introduce vision-specific inductive biases into the plain ViT, which can improve its performance on dense prediction tasks \citep{chenVisionTransformerAdapter2022}.\\
		Empirical studies have shown that Vision Transformers require more training data to perform similarly to convolutional neural networks. This is because Vision Transformers are data-hungry \citep{dosovitskiyImageWorth16x162021b}. To overcome this issue, researchers have proposed self-supervised pre-training. For example, one study introduced a self-supervised vision representation model called BEIT \citep{baoBEiTBERTPreTraining2022}.\\
		For segmentation, an article proposed a new architecture named Mask2Former that can address any segmentation tasks such as panoptic, instance, or semantic \citep{chengMaskedattentionMaskTransformer2022}.\\
		There are other segmentation architectures. For example, DeepLabv3+ is a semantic segmentation architecture based on the Xception network and Atrous Convolution. Arous convolution increases the resolution of feature maps, and spatial pyramid pooling is used to aggregate context information from multiple scales citep{chenEncoderDecoderAtrousSeparable2018}.\\
		Also, Segformer is a transformer-based architecture for semantic segmentation. Unlike traditional CNNs, it replaces convolutional operations with self-attention mechanisms. By doing so, the network can better capture long-range dependencies and contextual information in the input image citep{xieSegFormerSimpleEfficient2021}.\\
		Another advanced architecture for segmentation is Mask R-CNN. Mask R-CNN is a two-stage object detection and instance segmentation architecture. It extends Faster R-CNN by adding a branch for predicting an object mask and the existing branch for bounding box recognition. An end-to-end training process generates object proposals and predicts object masks citep{heMaskRCNN2018}.\\

		\subsection{Related works}
		Deep CNN algorithms were created to detect clinical dental periapical radiograph deterioration, periapical periodontitis, and periodontal disorders of mild, moderate, and high severity. The CNN model was used to explore classification, feature detection, segmentation, and quantification in periapical radiographs \citep{youDeepLearningbasedDental2020, liuSmartDentalHealthIoT2020a}.\\
		Also, studies have proven CNNs to detect pathological states in radiographs obtained in dental settings. These studies concentrated on detecting radiographic symptoms of maxillary sinusitis in panoramic radiographs and other types of radiographs commonly used in dentistry diagnosis and treatment \citep{kimDeepLearningDiagnosis2019,  murataDeeplearningClassificationUsing2019}.\\
		Segmentation algorithms have been used in different types of medical images. For example, the authors of one study proposed utilizing a spatially constrained convolutional neural network (SC-CNN) to detect and classify nuclei in histological pictures of common colon cancer in reference \citep{sirinukunwattanaLocalitySensitiveDeep2016}. Others proposed utilizing a U-net convolutional network to segment images from pulmonary CT to create a lung cancer screening system \citep{aitskourtLungCTImage2018}.\\
		Teeth segmentation has been the subject of many research projects in dental radiography. For instance, the seam carving technique includes preprocessing the X-ray images using adaptive thresholding before applying the segmentation algorithm \citep{al-sherifNewApproachTeeth2012}. Another study proposed a semi-automatic segmentation method for panoramic images in semi-automatic dental recognition. The proposed algorithm utilizes the Differential Image Foresting Transform (DIFT) to extract teeth' contours \citep{barbozaMultibiometricApproachSemi2012}. A research study has proposed using a deep learning method for separating and identifying each tooth in panoramic X-ray images \citep{jaderDeepInstanceSegmentation2018c}.\\
		However, it would be beneficial to introduce an architecture that can detect and identify teeth, roots, and multiple dental diseases and abnormalities such as pulp chamber, restoration, endodontics, crown, decay, pin, composite, bridge, pulpitis, orthodontics, radicular cyst, periapical cyst, cyst,  implant, and bone graft material in panoramic, periapical, and bitewing X-ray images. Therefore, this study proposes a combined methodology to identify all these structures and abnormalities (33 features) with better performance than the previous architectures. The rest of this article introduces the new architecture structure in the method section and compares the architecture’s performance with the earlier architectures in the result section.\\

		\section{Materials and Methods}
		\subsection{Dataset Collection}
		The data containing 963  OPG X-rays, 514  periapical X-rays, and 3673  bitewing X-rays were gathered from Valiasr Hospital in Tehran. We used these images for pre-training, and among these images, the number of 466  images were annotated with a group of 3 dentists. The annotated OPG images were manually split into the train and test groups to have different difficulty levels in both the training test groups, with a ratio of 90\% and 10\% in the training and test groups, respectively. All other X-rays, including periapical and bitewing, were added to the test group to assess the system's generalization. In pre-training, images were resized to (224 224), and for training, they were resized to (2048, 640).\\
		
		\subsection{Image Augmentation}
\begin{figure}
			
			\begin{center}
				\includegraphics[width=12.0cm,height=7.5cm]{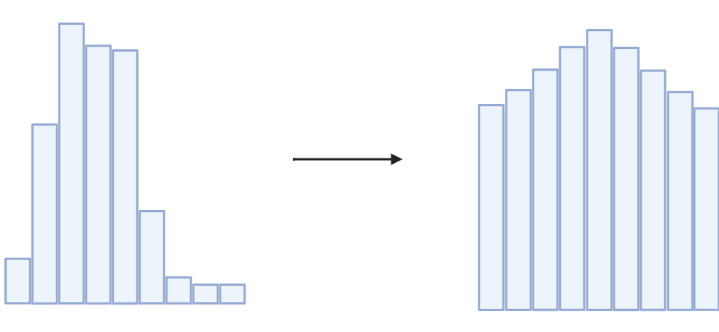}
				
				\caption{An Example of using Uniform Distributed Augmentation}
				\label{fig:uni}
				
			\end{center}
			
		\end{figure}

		For data augmentation, we used a new approach named Uniform Distributed Augmentation, which the number of images with smaller numbers increased more than images with larger numbers to have near to uniform distributions of images for better generalizations. An example of the process is shown in \ref{fig:uni}. The total number of augmented images was approximately 23000. Furthermore, data augmented occurred with a mathematical algorithm based on trial and error in implementing the best augmentation algorithm for current data. The mathematical process is:
		\[ f' = log(b + a * f) \]

		\subsection{Architecture Details}
\begin{figure}
			
			\begin{center}
				\includegraphics[width=14.5cm,height=6cm]{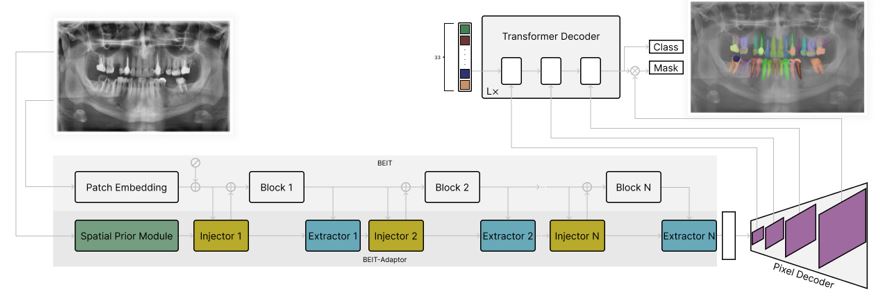}
				
				\caption{Overview of the proposed object detection architecture: The backbone consists of 24 transformer-based blocks with injectors and extractors. Preprocessing with BEIT is applied, and a Mask2Former decoder is used for object detection. Using a multi-scale decoding technique, the decoder produces OPG images with detected classes.}
				\label{fig:Capture}
				
			\end{center}
			
		\end{figure}

		Our method aims to propose a unique architecture for detecting and identifying teeth, dental diseases, and treated teeth, including pulp chamber, restoration, endodontics, crown, decay, pin, composite, bridge, pulpitis, orthodontics, radicular cyst, periapical cyst, cyst,  implant, and bone graft material in panoramic, periapical, and bitewing X-ray images.\\
		We implemented a Mask2Former \citep{chengMaskedattentionMaskTransformer2022a} decoder based on the vision transformer \citep{chenVisionTransformerAdapter2022a} and preprocessing with BEIT \citep{baoBEiTBERTPreTraining2022a}. The backbone consists of 24 transformer-based blocks with five injections and extractions. The extracted representation of the last transformer block is fed into the Mask2Former decoder. By providing one scale of the multi-scale feature to one Transformer decoder layer at a time, the decoder effectively utilizes high-resolution features from a pixel decoder \citep{chengMaskedattentionMaskTransformer2022a}.\\
		In the pre-training phase, 1300 augmented OPG images with Uniform distribution Augmentation are fed into the BEIT encoder. BEIT pre-training proposed a masked image modeling task that employs two image views; image patches and visual tokens. Some image patches are randomly masked and replaced with a special mask embedding. After that, the image patches are fed to a backbone vision transformer. The pre-training seeks to predict the visual tokens of the original picture based on the corrupted image's encoding vectors \citep{baoBEiTBERTPreTraining2022a}.\\
		In BEIT training, the input images are first fed into 24 transformer blocks. After the pre-training, the transformer blocks are connected to injectors and extractors in BEIT-Adaptor. In BEIT-Adaptor, the spatial prior module models local spatial contexts from the input image, the spatial feature injector introduces spatial priors into the BEIT, and the multi-scale feature extractor reconstructs multi-scale features from BEITS's single-scale features.\\
		In the decoder part, Mask2Former \citep{chengMaskedattentionMaskTransformer2022a} 
		is used. In Mask2Former, instead of attending to the whole feature map, the transformer decoder includes a masked attention operator that extracts localized features by constraining mask attention inside the foreground area of the predicted mask for each query. A multi-scale technique is implemented to handle small objects that use high-resolution features. In the transformer decoder, 33 features are fed into the network, and the outputs are OPG images with detected classes.  The architecture overview is shown in Figure \ref{fig:Capture}.\\
		
		\section{Result}
\begin{figure}
			\centering
			
			\subfigure[]{%
				\includegraphics[width=6.5cm,height=6cm]{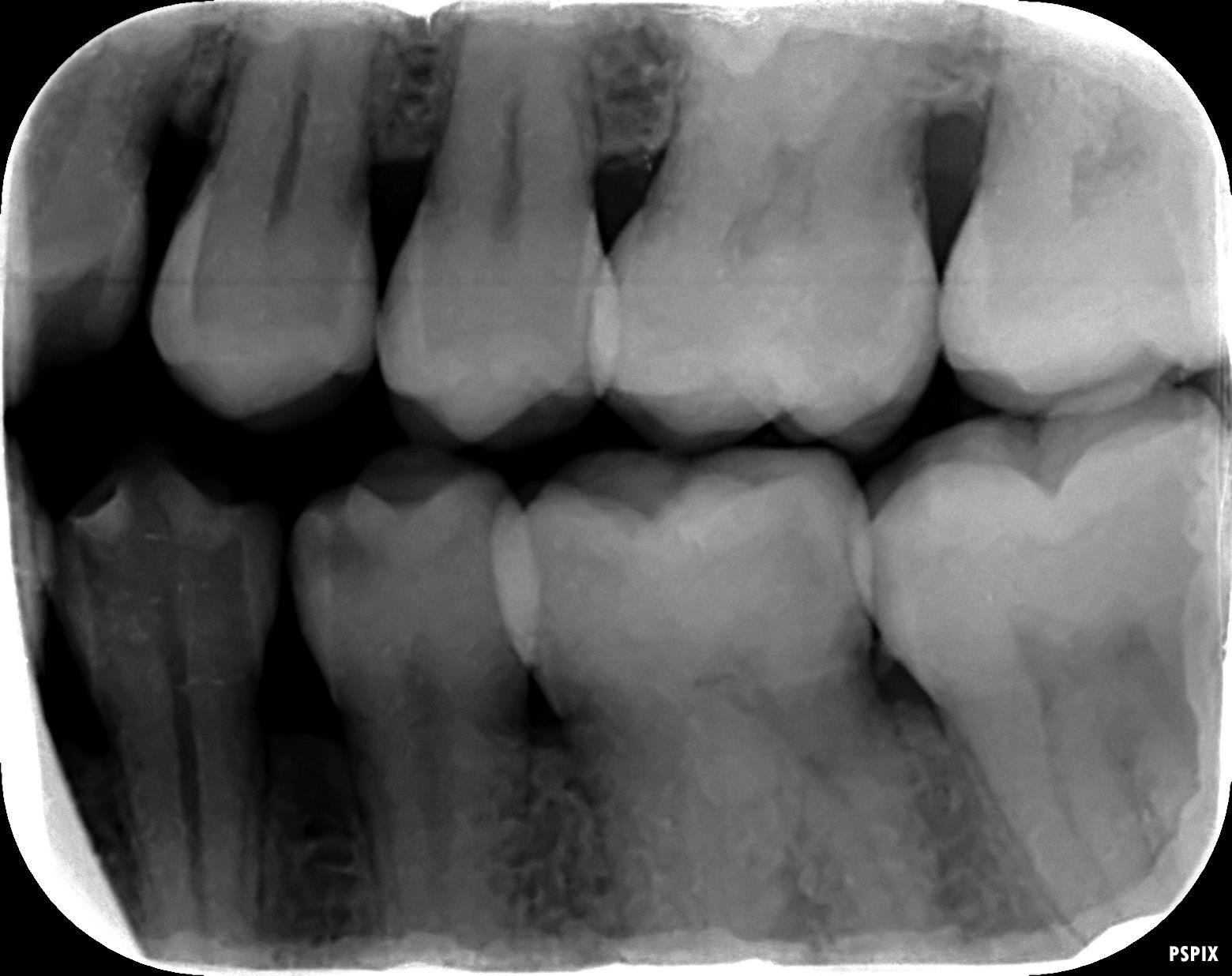}
				\label{fig:1}}
			\hfill
			\subfigure[]{%
				\includegraphics[width=6.5cm,height=6cm]{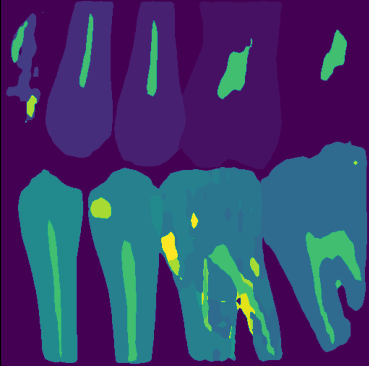}
				\label{fig:2}}
			
			\subfigure[]{%
				\includegraphics[width=6.5cm,height=6cm]{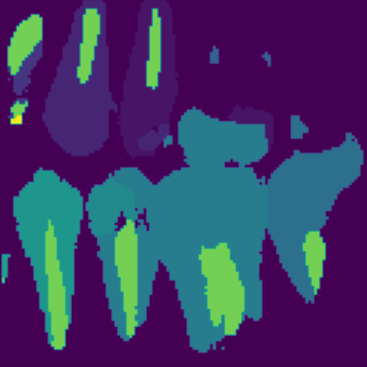}
				\label{fig:3}}
			\hfill
			\subfigure[]{%
				\includegraphics[width=6.5cm,height=6cm]{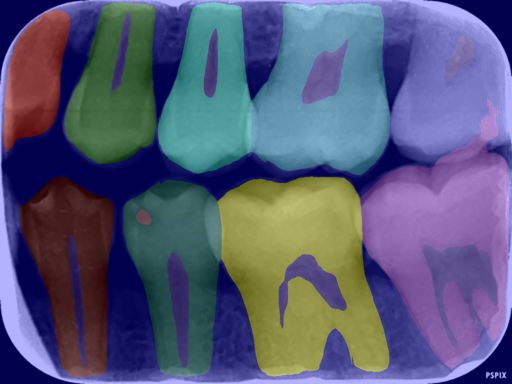}
				\label{fig:4}}
			
			\caption{This figure illustrates (a) periapical X-ray and detected outputs of this image using (b) Deeplabv3+, (c) Segformer, and (d) Radious}
			\label{fig:all}
		\end{figure}

		In this part, we describe the findings of our proposed new architecture for segmenting dental X-ray images. We aimed to do semantic segmentation of dental X-ray images employing our new architecture, which consists of BEIT-Adaptor and Mask2Former techniques. We also utilize our Uniform Distributed Augmentation technique for training. We tested our model's efficiency using two common segmentation performance metrics: mean Intersection over Union (mIoU) and mean Accuracy (mAcc). In addition, we compared our architecture's performance to that of two popular and cutting-edge segmentation models: DeepLabv3+ and Segformer \citep{chenEncoderDecoderAtrousSeparable2018a, xieSegFormerSimpleEfficient2021}.
		In order to evaluate the performance of image segmentation algorithms, the mIoU metric is commonly used. The intersection area of the predicted segmentation mask and the ground truth mask is compared to their union area to calculate the ratio. The mIoU score is obtained by averaging the IoU values for each class in the sample. As a result of a better match between the predicted and ground truth masks, a higher mIoU value indicates better segmentation performance.\\
		Another metric used to assess segmentation performance is mAcc. It calculates the proportion of correctly categorized pixels in relation to the total number of pixels in the image. The average of per-class accuracies across all classes in the dataset calculates mAcc. A higher mAcc score denotes greater segmentation performance since it represents a higher percentage of correctly classified pixels.\\
		Our proposed architecture received a mIoU score of 90\% and a mAcc score of 65\% in our tests. These findings indicate that our model can accurately segment dental X-ray images.\\
		To further validate the performance of our proposed architecture, we compared it with the results of DeepLabv3+ and Segformer. DeepLabv3+ obtained a mIoU of 85.7\% in its respective article \citep{chenEncoderDecoderAtrousSeparable2018a}, whereas Segformer earned a mIoU of 83.1\% \citep{xieSegFormerSimpleEfficient2021} with their own dataset.\\
		In addition, we evaluated the performance of our architecture in comparison to DeepLabv3+ and Segformer models with our own dataset. Compared to DeepLabv3+ and Segformer, our model outperformed existing implementations, raising the mIoU score by 9 and 33 percentage points, respectively. The results are shown in Table ~\ref{tab:table1},  and sample visualization of detected X-rays are shown in Figure \ref{fig:all}\\
		
\begin{figure}
			
			\begin{center}
				\includegraphics[width=14.5cm,height=8cm]{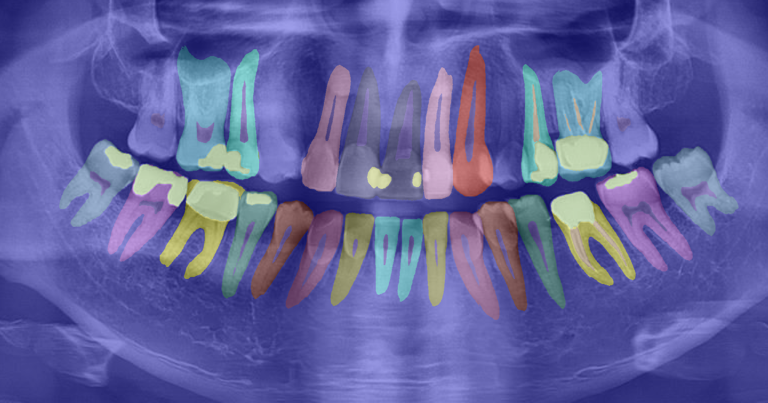}
				
				\caption{The dental radiographic output image The objects (classes) are exist in radiograph. image: upper second premolar, endo, lower lateral, lower second molar, lower third molar, restoration, upper second molar, lower first molar, upper third molar, lower first premolar, upper first molar, upper central, lower second premolar, lower canine, composite, lower central, upper canine, upper first premolar, pulp chamber}
				\label{fig:7}
				
			\end{center}
			
		\end{figure}

		These results demonstrate our suggested architecture's improved semantic segmentation performance on dental X-ray images. A sample detected dental X-ray (OPG) is shown in Figure \ref{fig:7}.\\
		
		\begin{table}
			\begin{center}
				\caption{Comparing mIoU and mAcc scores among our algorithm (Radious), Deeplabv3+ and Segformer}
				\label{tab:table1}
				\begin{tabular}{@{}llr@{}} 
					\toprule
					\textbf{Algorithm} & \textbf{mIoU} & \textbf{mAcc} \\ \midrule
					Deeplabv3+ & 0.56 & 0.70 \\
					Segformer & 0.32 & 0.15 \\
					Radious & 0.65 & 0.90 \\
					\bottomrule
				\end{tabular}
			\end{center}
		\end{table}

		\section{Discussion}
		Our study focused on developing a unique architecture for dental X-ray image segmentation to outperform existing state-of-the-art models such as DeepLabv3+ and Segformer regarding mIoU and mAcc scores. Our suggested design outperformed DeepLabv3+ and Segformer regarding mIoU and mAcc scores in our dataset, confirming our initial hypothesis. The improvements in segmentation performance are due to the novel architecture, which consists of BEIT-Adaptor as an encoder and Mask2Former as a decoder.\\
		One notable aspect of our work was the use of 33 labels, including teeth, roots, and multiple dental diseases and abnormalities such as pulp chamber, restoration, endodontics, crown, decay, pin, composite, bridge, pulpitis, orthodontics, radicular cyst, periapical cyst, cyst, implant, and bone graft, which is significantly higher than the number of labels used in previous studies to the best of our knowledge. With additional labels, our model could more accurately represent the complex and various structures found in dental X-ray images. Consequently, in practical situations, our approach demonstrated improved accuracy and resiliency.\\
		The Uniform Distributed Augmentation data augmentation method we used is another notable accomplishment of our study. With a near-uniform distribution of images for better generalization, this method specifically aimed to increase the number of images with smaller sample sizes more than those with larger sample sizes. The effective use of this method demonstrates its capability to improve segmentation model performance in situations with sparse or unbalanced data.\\
		Despite the encouraging findings, our study had some drawbacks. One such restriction was the number of images we used for training and validation. As previously discussed, we used augmentation techniques to increase our dataset's diversity and size. However, the performance and generalizability of the model might be improved by using a larger dataset with a greater number of dental X-ray images. Another limitation was the difficulty of labeling the dental X-ray images, which required professional expertise. Labeling is so time-consuming and hard work.\\ 
		For future work, we advise improving the model for classifying additional dental conditions, such as various cysts, various levels of caries, jaw arch analysis, and bone recession assessment for surgical planning, which would significantly advance dental diagnosis and treatment planning. Finally, exploring the potential of our architecture for 3D dental imaging techniques like cone-beam computed tomography (CBCT) may increase its application and impact in the dental field.\\

		\bibliographystyle{elsarticle-harv}
		\bibliography{MyLibrary}

\begin{thebibliography}{27}
\expandafter\ifx\csname natexlab\endcsname\relax\def\natexlab#1{#1}\fi
\providecommand{\url}[1]{\texttt{#1}}
\providecommand{\href}[2]{#2}
\providecommand{\path}[1]{#1}
\providecommand{\DOIprefix}{doi:}
\providecommand{\ArXivprefix}{arXiv:}
\providecommand{\URLprefix}{URL: }
\providecommand{\Pubmedprefix}{pmid:}
\providecommand{\doi}[1]{\href{http://dx.doi.org/#1}{\path{#1}}}
\providecommand{\Pubmed}[1]{\href{pmid:#1}{\path{#1}}}
\providecommand{\bibinfo}[2]{#2}
\ifx\xfnm\relax \def\xfnm[#1]{\unskip,\space#1}\fi
\bibitem[{Ait~Skourt et~al.(2018)Ait~Skourt, El~Hassani and
  Majda}]{aitskourtLungCTImage2018}
\bibinfo{author}{Ait~Skourt, B.}, \bibinfo{author}{El~Hassani, A.},
  \bibinfo{author}{Majda, A.}, \bibinfo{year}{2018}.
\newblock \bibinfo{title}{Lung {{CT Image Segmentation Using Deep Neural
  Networks}}}.
\newblock \bibinfo{journal}{Procedia Computer Science} \bibinfo{volume}{127},
  \bibinfo{pages}{109--113}.
\newblock \DOIprefix\doi{10.1016/j.procs.2018.01.104}.
\bibitem[{{Al-sherif} et~al.(2012){Al-sherif}, Guo and
  Ammar}]{al-sherifNewApproachTeeth2012}
\bibinfo{author}{{Al-sherif}, N.}, \bibinfo{author}{Guo, G.},
  \bibinfo{author}{Ammar, H.}, \bibinfo{year}{2012}.
\newblock \bibinfo{title}{A {{New Approach}} to {{Teeth Segmentation}}}.
\newblock \DOIprefix\doi{10.1109/ISM.2012.35}.
\bibitem[{Bailit et~al.(1980)Bailit, Reisine, Damuth and
  Richards}]{bailitValidityRadiographicMethod1980a}
\bibinfo{author}{Bailit, H.L.}, \bibinfo{author}{Reisine, S.T.},
  \bibinfo{author}{Damuth, R.L.}, \bibinfo{author}{Richards, N.P.},
  \bibinfo{year}{1980}.
\newblock \bibinfo{title}{The validity of the radiographic method in the
  pretreatment review of dental claims}.
\newblock \bibinfo{journal}{Journal of Public Health Dentistry}
  \bibinfo{volume}{40}, \bibinfo{pages}{26--38}.
\newblock \DOIprefix\doi{10.1111/j.1752-7325.1980.tb01846.x}.
\bibitem[{Bao et~al.(2022a)Bao, Dong, Piao and
  Wei}]{baoBEiTBERTPreTraining2022}
\bibinfo{author}{Bao, H.}, \bibinfo{author}{Dong, L.}, \bibinfo{author}{Piao,
  S.}, \bibinfo{author}{Wei, F.}, \bibinfo{year}{2022}a.
\newblock \bibinfo{title}{{{BEiT}}: {{BERT Pre-Training}} of {{Image
  Transformers}}}.
\newblock \href{http://arxiv.org/abs/2106.08254}{{\tt arXiv:2106.08254}}.
\bibitem[{Bao et~al.(2022b)Bao, Dong, Piao and
  Wei}]{baoBEiTBERTPreTraining2022a}
\bibinfo{author}{Bao, H.}, \bibinfo{author}{Dong, L.}, \bibinfo{author}{Piao,
  S.}, \bibinfo{author}{Wei, F.}, \bibinfo{year}{2022}b.
\newblock \bibinfo{title}{{{BEiT}}: {{BERT Pre-Training}} of {{Image
  Transformers}}}.
\newblock \href{http://arxiv.org/abs/2106.08254}{{\tt arXiv:2106.08254}}.
\bibitem[{Barboza(2012)}]{barbozaMultibiometricApproachSemi2012}
\bibinfo{author}{Barboza, E.B.}, \bibinfo{year}{2012}.
\newblock \bibinfo{title}{A {{Multibiometric Approach}} in a {{Semi Automatic
  Dental Recognition Using DIFT Technique}} and {{Dental Shape Features}}}.
\bibitem[{Chen et~al.(2018)Chen, Zhu, Papandreou, Schroff and
  Adam}]{chenEncoderDecoderAtrousSeparable2018a}
\bibinfo{author}{Chen, L.C.}, \bibinfo{author}{Zhu, Y.},
  \bibinfo{author}{Papandreou, G.}, \bibinfo{author}{Schroff, F.},
  \bibinfo{author}{Adam, H.}, \bibinfo{year}{2018}.
\newblock \bibinfo{title}{Encoder-{{Decoder}} with {{Atrous Separable
  Convolution}} for {{Semantic Image Segmentation}}}.
\newblock \href{http://arxiv.org/abs/1802.02611}{{\tt arXiv:1802.02611}}.
\bibitem[{Chen et~al.(2022a)Chen, Duan, Wang, He, Lu, Dai and
  Qiao}]{chenVisionTransformerAdapter2022}
\bibinfo{author}{Chen, Z.}, \bibinfo{author}{Duan, Y.}, \bibinfo{author}{Wang,
  W.}, \bibinfo{author}{He, J.}, \bibinfo{author}{Lu, T.},
  \bibinfo{author}{Dai, J.}, \bibinfo{author}{Qiao, Y.}, \bibinfo{year}{2022}a.
\newblock \bibinfo{title}{Vision {{Transformer Adapter}} for {{Dense
  Predictions}}}.
\newblock \href{http://arxiv.org/abs/2205.08534}{{\tt arXiv:2205.08534}}.
\bibitem[{Chen et~al.(2022b)Chen, Duan, Wang, He, Lu, Dai and
  Qiao}]{chenVisionTransformerAdapter2022a}
\bibinfo{author}{Chen, Z.}, \bibinfo{author}{Duan, Y.}, \bibinfo{author}{Wang,
  W.}, \bibinfo{author}{He, J.}, \bibinfo{author}{Lu, T.},
  \bibinfo{author}{Dai, J.}, \bibinfo{author}{Qiao, Y.}, \bibinfo{year}{2022}b.
\newblock \bibinfo{title}{Vision {{Transformer Adapter}} for {{Dense
  Predictions}}}.
\newblock \href{http://arxiv.org/abs/2205.08534}{{\tt arXiv:2205.08534}}.
\bibitem[{Cheng et~al.(2022a)Cheng, Misra, Schwing, Kirillov and
  Girdhar}]{chengMaskedattentionMaskTransformer2022}
\bibinfo{author}{Cheng, B.}, \bibinfo{author}{Misra, I.},
  \bibinfo{author}{Schwing, A.G.}, \bibinfo{author}{Kirillov, A.},
  \bibinfo{author}{Girdhar, R.}, \bibinfo{year}{2022}a.
\newblock \bibinfo{title}{Masked-attention {{Mask Transformer}} for {{Universal
  Image Segmentation}}}.
\newblock \href{http://arxiv.org/abs/2112.01527}{{\tt arXiv:2112.01527}}.
\bibitem[{Cheng et~al.(2022b)Cheng, Misra, Schwing, Kirillov and
  Girdhar}]{chengMaskedattentionMaskTransformer2022a}
\bibinfo{author}{Cheng, B.}, \bibinfo{author}{Misra, I.},
  \bibinfo{author}{Schwing, A.G.}, \bibinfo{author}{Kirillov, A.},
  \bibinfo{author}{Girdhar, R.}, \bibinfo{year}{2022}b.
\newblock \bibinfo{title}{Masked-attention {{Mask Transformer}} for {{Universal
  Image Segmentation}}}.
\newblock \href{http://arxiv.org/abs/2112.01527}{{\tt arXiv:2112.01527}}.
\bibitem[{Dosovitskiy et~al.(2021a)Dosovitskiy, Beyer, Kolesnikov, Weissenborn,
  Zhai, Unterthiner, Dehghani, Minderer, Heigold, Gelly, Uszkoreit and
  Houlsby}]{dosovitskiyImageWorth16x162021a}
\bibinfo{author}{Dosovitskiy, A.}, \bibinfo{author}{Beyer, L.},
  \bibinfo{author}{Kolesnikov, A.}, \bibinfo{author}{Weissenborn, D.},
  \bibinfo{author}{Zhai, X.}, \bibinfo{author}{Unterthiner, T.},
  \bibinfo{author}{Dehghani, M.}, \bibinfo{author}{Minderer, M.},
  \bibinfo{author}{Heigold, G.}, \bibinfo{author}{Gelly, S.},
  \bibinfo{author}{Uszkoreit, J.}, \bibinfo{author}{Houlsby, N.},
  \bibinfo{year}{2021}a.
\newblock \bibinfo{title}{An {{Image}} is {{Worth}} 16x16 {{Words}}:
  {{Transformers}} for {{Image Recognition}} at {{Scale}}}.
\newblock \DOIprefix\doi{10.48550/arXiv.2010.11929},
  \href{http://arxiv.org/abs/2010.11929}{{\tt arXiv:2010.11929}}.
\bibitem[{Dosovitskiy et~al.(2021b)Dosovitskiy, Beyer, Kolesnikov, Weissenborn,
  Zhai, Unterthiner, Dehghani, Minderer, Heigold, Gelly, Uszkoreit and
  Houlsby}]{dosovitskiyImageWorth16x162021}
\bibinfo{author}{Dosovitskiy, A.}, \bibinfo{author}{Beyer, L.},
  \bibinfo{author}{Kolesnikov, A.}, \bibinfo{author}{Weissenborn, D.},
  \bibinfo{author}{Zhai, X.}, \bibinfo{author}{Unterthiner, T.},
  \bibinfo{author}{Dehghani, M.}, \bibinfo{author}{Minderer, M.},
  \bibinfo{author}{Heigold, G.}, \bibinfo{author}{Gelly, S.},
  \bibinfo{author}{Uszkoreit, J.}, \bibinfo{author}{Houlsby, N.},
  \bibinfo{year}{2021}b.
\newblock \bibinfo{title}{An {{Image}} is {{Worth}} 16x16 {{Words}}:
  {{Transformers}} for {{Image Recognition}} at {{Scale}}}.
\newblock \DOIprefix\doi{10.48550/arXiv.2010.11929},
  \href{http://arxiv.org/abs/2010.11929}{{\tt arXiv:2010.11929}}.
\bibitem[{Dosovitskiy et~al.(2021c)Dosovitskiy, Beyer, Kolesnikov, Weissenborn,
  Zhai, Unterthiner, Dehghani, Minderer, Heigold, Gelly, Uszkoreit and
  Houlsby}]{dosovitskiyImageWorth16x162021b}
\bibinfo{author}{Dosovitskiy, A.}, \bibinfo{author}{Beyer, L.},
  \bibinfo{author}{Kolesnikov, A.}, \bibinfo{author}{Weissenborn, D.},
  \bibinfo{author}{Zhai, X.}, \bibinfo{author}{Unterthiner, T.},
  \bibinfo{author}{Dehghani, M.}, \bibinfo{author}{Minderer, M.},
  \bibinfo{author}{Heigold, G.}, \bibinfo{author}{Gelly, S.},
  \bibinfo{author}{Uszkoreit, J.}, \bibinfo{author}{Houlsby, N.},
  \bibinfo{year}{2021}c.
\newblock \bibinfo{title}{An {{Image}} is {{Worth}} 16x16 {{Words}}:
  {{Transformers}} for {{Image Recognition}} at {{Scale}}}.
\newblock \DOIprefix\doi{10.48550/arXiv.2010.11929},
  \href{http://arxiv.org/abs/2010.11929}{{\tt arXiv:2010.11929}}.
\bibitem[{Huang et~al.(2021)Huang, Ben, Luo, Cheng, Yu and
  Fu}]{huangShuffleTransformerRethinking2021}
\bibinfo{author}{Huang, Z.}, \bibinfo{author}{Ben, Y.}, \bibinfo{author}{Luo,
  G.}, \bibinfo{author}{Cheng, P.}, \bibinfo{author}{Yu, G.},
  \bibinfo{author}{Fu, B.}, \bibinfo{year}{2021}.
\newblock \bibinfo{title}{Shuffle {{Transformer}}: {{Rethinking Spatial
  Shuffle}} for {{Vision Transformer}}}.
\newblock \DOIprefix\doi{10.48550/arXiv.2106.03650},
  \href{http://arxiv.org/abs/2106.03650}{{\tt arXiv:2106.03650}}.
\bibitem[{Jader et~al.(2018)Jader, Fontineli, Ruiz, Abdalla, Pithon and
  Oliveira}]{jaderDeepInstanceSegmentation2018c}
\bibinfo{author}{Jader, G.}, \bibinfo{author}{Fontineli, J.},
  \bibinfo{author}{Ruiz, M.}, \bibinfo{author}{Abdalla, K.},
  \bibinfo{author}{Pithon, M.}, \bibinfo{author}{Oliveira, L.},
  \bibinfo{year}{2018}.
\newblock \bibinfo{title}{Deep {{Instance Segmentation}} of {{Teeth}} in
  {{Panoramic X-Ray Images}}}, in: \bibinfo{booktitle}{2018 31st {{SIBGRAPI
  Conference}} on {{Graphics}}, {{Patterns}} and {{Images}} ({{SIBGRAPI}})},
  \bibinfo{publisher}{{IEEE}}, \bibinfo{address}{{Parana}}. pp.
  \bibinfo{pages}{400--407}.
\newblock \DOIprefix\doi{10.1109/SIBGRAPI.2018.00058}.
\bibitem[{Kermany et~al.(2018)Kermany, Goldbaum, Cai, Valentim, Liang, Baxter,
  McKeown, Yang, Wu, Yan, Dong, Prasadha, Pei, Ting, Zhu, Li, Hewett, Dong,
  Ziyar, Shi, Zhang, Zheng, Hou, Shi, Fu, Duan, Huu, Wen, Zhang, Zhang, Li,
  Wang, Singer, Sun, Xu, Tafreshi, Lewis, Xia and
  Zhang}]{kermanyIdentifyingMedicalDiagnoses2018}
\bibinfo{author}{Kermany, D.S.}, \bibinfo{author}{Goldbaum, M.},
  \bibinfo{author}{Cai, W.}, \bibinfo{author}{Valentim, C.C.S.},
  \bibinfo{author}{Liang, H.}, \bibinfo{author}{Baxter, S.L.},
  \bibinfo{author}{McKeown, A.}, \bibinfo{author}{Yang, G.},
  \bibinfo{author}{Wu, X.}, \bibinfo{author}{Yan, F.}, \bibinfo{author}{Dong,
  J.}, \bibinfo{author}{Prasadha, M.K.}, \bibinfo{author}{Pei, J.},
  \bibinfo{author}{Ting, M.Y.L.}, \bibinfo{author}{Zhu, J.},
  \bibinfo{author}{Li, C.}, \bibinfo{author}{Hewett, S.},
  \bibinfo{author}{Dong, J.}, \bibinfo{author}{Ziyar, I.},
  \bibinfo{author}{Shi, A.}, \bibinfo{author}{Zhang, R.},
  \bibinfo{author}{Zheng, L.}, \bibinfo{author}{Hou, R.}, \bibinfo{author}{Shi,
  W.}, \bibinfo{author}{Fu, X.}, \bibinfo{author}{Duan, Y.},
  \bibinfo{author}{Huu, V.A.N.}, \bibinfo{author}{Wen, C.},
  \bibinfo{author}{Zhang, E.D.}, \bibinfo{author}{Zhang, C.L.},
  \bibinfo{author}{Li, O.}, \bibinfo{author}{Wang, X.},
  \bibinfo{author}{Singer, M.A.}, \bibinfo{author}{Sun, X.},
  \bibinfo{author}{Xu, J.}, \bibinfo{author}{Tafreshi, A.},
  \bibinfo{author}{Lewis, M.A.}, \bibinfo{author}{Xia, H.},
  \bibinfo{author}{Zhang, K.}, \bibinfo{year}{2018}.
\newblock \bibinfo{title}{Identifying {{Medical Diagnoses}} and {{Treatable
  Diseases}} by {{Image-Based Deep Learning}}}.
\newblock \bibinfo{journal}{Cell} \bibinfo{volume}{172},
  \bibinfo{pages}{1122--1131.e9}.
\newblock \DOIprefix\doi{10.1016/j.cell.2018.02.010}.
\bibitem[{Kim et~al.(2019)Kim, Lee, Sunwoo, Choi, Nam, Cho, Kim, Bae, Yoo,
  Choi, Jung and Kim}]{kimDeepLearningDiagnosis2019}
\bibinfo{author}{Kim, Y.}, \bibinfo{author}{Lee, K.J.},
  \bibinfo{author}{Sunwoo, L.}, \bibinfo{author}{Choi, D.},
  \bibinfo{author}{Nam, C.M.}, \bibinfo{author}{Cho, J.}, \bibinfo{author}{Kim,
  J.}, \bibinfo{author}{Bae, Y.J.}, \bibinfo{author}{Yoo, R.E.},
  \bibinfo{author}{Choi, B.S.}, \bibinfo{author}{Jung, C.},
  \bibinfo{author}{Kim, J.H.}, \bibinfo{year}{2019}.
\newblock \bibinfo{title}{Deep {{Learning}} in {{Diagnosis}} of {{Maxillary
  Sinusitis Using Conventional Radiography}}}.
\newblock \bibinfo{journal}{Investigative Radiology} \bibinfo{volume}{54},
  \bibinfo{pages}{7--15}.
\newblock \DOIprefix\doi{10.1097/RLI.0000000000000503}.
\bibitem[{Langlais et~al.(1987)Langlais, Skoczylas, Prihoda, Langland and
  Schiff}]{langlaisInterpretationBitewingRadiographs1987b}
\bibinfo{author}{Langlais, R.P.}, \bibinfo{author}{Skoczylas, L.J.},
  \bibinfo{author}{Prihoda, T.J.}, \bibinfo{author}{Langland, O.E.},
  \bibinfo{author}{Schiff, T.}, \bibinfo{year}{1987}.
\newblock \bibinfo{title}{Interpretation of bitewing radiographs:
  {{Application}} of the kappa statistic to determine rater agreements}.
\newblock \bibinfo{journal}{Oral Surgery, Oral Medicine, Oral Pathology}
  \bibinfo{volume}{64}, \bibinfo{pages}{751--756}.
\newblock \DOIprefix\doi{10.1016/0030-4220(87)90181-2}.
\bibitem[{Liu et~al.(2020)Liu, Xu, Huan, Zou, Yeh and
  Zheng}]{liuSmartDentalHealthIoT2020a}
\bibinfo{author}{Liu, L.}, \bibinfo{author}{Xu, J.}, \bibinfo{author}{Huan,
  Y.}, \bibinfo{author}{Zou, Z.}, \bibinfo{author}{Yeh, S.C.},
  \bibinfo{author}{Zheng, L.R.}, \bibinfo{year}{2020}.
\newblock \bibinfo{title}{A {{Smart Dental Health-IoT Platform Based}} on
  {{Intelligent Hardware}}, {{Deep Learning}}, and {{Mobile Terminal}}}.
\newblock \bibinfo{journal}{IEEE Journal of Biomedical and Health Informatics}
  \bibinfo{volume}{24}, \bibinfo{pages}{898--906}.
\newblock \DOIprefix\doi{10.1109/JBHI.2019.2919916}.
\bibitem[{Long et~al.(2015)Long, Shelhamer and
  Darrell}]{longFullyConvolutionalNetworks2015}
\bibinfo{author}{Long, J.}, \bibinfo{author}{Shelhamer, E.},
  \bibinfo{author}{Darrell, T.}, \bibinfo{year}{2015}.
\newblock \bibinfo{title}{Fully {{Convolutional Networks}} for {{Semantic
  Segmentation}}}.
\newblock \DOIprefix\doi{10.48550/arXiv.1411.4038},
  \href{http://arxiv.org/abs/1411.4038}{{\tt arXiv:1411.4038}}.
\bibitem[{Murata et~al.(2019)Murata, Ariji, Ohashi, Kawai, Fukuda, Funakoshi,
  Kise, Nozawa, Katsumata, Fujita and
  Ariji}]{murataDeeplearningClassificationUsing2019}
\bibinfo{author}{Murata, M.}, \bibinfo{author}{Ariji, Y.},
  \bibinfo{author}{Ohashi, Y.}, \bibinfo{author}{Kawai, T.},
  \bibinfo{author}{Fukuda, M.}, \bibinfo{author}{Funakoshi, T.},
  \bibinfo{author}{Kise, Y.}, \bibinfo{author}{Nozawa, M.},
  \bibinfo{author}{Katsumata, A.}, \bibinfo{author}{Fujita, H.},
  \bibinfo{author}{Ariji, E.}, \bibinfo{year}{2019}.
\newblock \bibinfo{title}{Deep-learning classification using convolutional
  neural network for evaluation of maxillary sinusitis on panoramic
  radiography}.
\newblock \bibinfo{journal}{Oral Radiology} \bibinfo{volume}{35},
  \bibinfo{pages}{301--307}.
\newblock \DOIprefix\doi{10.1007/s11282-018-0363-7}.
\bibitem[{Sirinukunwattana et~al.(2016)Sirinukunwattana, Ahmed~Raza, {Yee-Wah
  Tsang}, Snead, Cree and Rajpoot}]{sirinukunwattanaLocalitySensitiveDeep2016}
\bibinfo{author}{Sirinukunwattana, K.}, \bibinfo{author}{Ahmed~Raza, S.E.},
  \bibinfo{author}{{Yee-Wah Tsang}, n.}, \bibinfo{author}{Snead, D.R.J.},
  \bibinfo{author}{Cree, I.A.}, \bibinfo{author}{Rajpoot, N.M.},
  \bibinfo{year}{2016}.
\newblock \bibinfo{title}{Locality {{Sensitive Deep Learning}} for
  {{Detection}} and {{Classification}} of {{Nuclei}} in {{Routine Colon Cancer
  Histology Images}}}.
\newblock \bibinfo{journal}{IEEE transactions on medical imaging}
  \bibinfo{volume}{35}, \bibinfo{pages}{1196--1206}.
\newblock \DOIprefix\doi{10.1109/TMI.2016.2525803}.
\bibitem[{Strudel et~al.(2021)Strudel, Garcia, Laptev and
  Schmid}]{strudelSegmenterTransformerSemantic2021}
\bibinfo{author}{Strudel, R.}, \bibinfo{author}{Garcia, R.},
  \bibinfo{author}{Laptev, I.}, \bibinfo{author}{Schmid, C.},
  \bibinfo{year}{2021}.
\newblock \bibinfo{title}{Segmenter: {{Transformer}} for {{Semantic
  Segmentation}}}.
\newblock \href{http://arxiv.org/abs/2105.05633}{{\tt arXiv:2105.05633}}.
\bibitem[{Wang et~al.(2016)Wang, Huang, Lee, Li, Chang, Siao, Lai, Ibragimov,
  Vrtovec, Ronneberger, Fischer, Cootes and
  Lindner}]{wangBenchmarkComparisonDental2016}
\bibinfo{author}{Wang, C.W.}, \bibinfo{author}{Huang, C.T.},
  \bibinfo{author}{Lee, J.H.}, \bibinfo{author}{Li, C.H.},
  \bibinfo{author}{Chang, S.W.}, \bibinfo{author}{Siao, M.J.},
  \bibinfo{author}{Lai, T.M.}, \bibinfo{author}{Ibragimov, B.},
  \bibinfo{author}{Vrtovec, T.}, \bibinfo{author}{Ronneberger, O.},
  \bibinfo{author}{Fischer, P.}, \bibinfo{author}{Cootes, T.F.},
  \bibinfo{author}{Lindner, C.}, \bibinfo{year}{2016}.
\newblock \bibinfo{title}{A benchmark for comparison of dental radiography
  analysis algorithms}.
\newblock \bibinfo{journal}{Medical Image Analysis} \bibinfo{volume}{31},
  \bibinfo{pages}{63--76}.
\newblock \DOIprefix\doi{10.1016/j.media.2016.02.004}.
\bibitem[{Xie et~al.(2021)Xie, Wang, Yu, Anandkumar, Alvarez and
  Luo}]{xieSegFormerSimpleEfficient2021}
\bibinfo{author}{Xie, E.}, \bibinfo{author}{Wang, W.}, \bibinfo{author}{Yu,
  Z.}, \bibinfo{author}{Anandkumar, A.}, \bibinfo{author}{Alvarez, J.M.},
  \bibinfo{author}{Luo, P.}, \bibinfo{year}{2021}.
\newblock \bibinfo{title}{{{SegFormer}}: {{Simple}} and {{Efficient Design}}
  for {{Semantic Segmentation}} with {{Transformers}}}.
\newblock \DOIprefix\doi{10.48550/arXiv.2105.15203},
  \href{http://arxiv.org/abs/2105.15203}{{\tt arXiv:2105.15203}}.
\bibitem[{You et~al.(2020)You, Hao, Li, Wang and
  Xia}]{youDeepLearningbasedDental2020}
\bibinfo{author}{You, W.}, \bibinfo{author}{Hao, A.}, \bibinfo{author}{Li, S.},
  \bibinfo{author}{Wang, Y.}, \bibinfo{author}{Xia, B.}, \bibinfo{year}{2020}.
\newblock \bibinfo{title}{Deep learning-based dental plaque detection on
  primary teeth: A comparison with clinical assessments}.
\newblock \bibinfo{journal}{BMC Oral Health} \bibinfo{volume}{20},
  \bibinfo{pages}{141}.
\newblock \DOIprefix\doi{10.1186/s12903-020-01114-6}.

\end{thebibliography}
		
		
			
			
			
	\end{document}